# Application of Freeman Chain Codes: An Alternative Recognition Technique for Malaysian Car Plates

***Nor Amizam Jusoh*** [†] **and** ***Jasni Mohamad Zain*** [††],

***[†]IKIP International College, Taman Gelora, 25000 Kuantan, Pahang Malaysia***
***Faculty of Computer Systems & Software Engineering,***
***[††]University Malaysia Pahang, Locked Bag 12, 25000 Kuantan, Pahang Malaysia***

**Summary**
*Various applications of car plate recognition systems have been developed using various kinds of methods and techniques by researchers all over the world. The applications developed were only suitable for specific country due to its standard specification endorsed by the transport department of particular countries. The Road Transport Department of Malaysia also has endorsed a specification for car plates that includes the font and size of characters that must be followed by car owners. However, there are cases where this specification is not followed. Several applications have been developed in Malaysia to overcome this problem. However, there is still problem in achieving 100% recognition accuracy. This paper is mainly focused on conducting an experiment using chain codes technique to perform recognition for different types of fonts used in Malaysian car plates.*
***Keywords***
*Image processing, recognition, chain codes, segmentation*

## 1. Introduction

Car plate recognition system is a complex image processing application which recognizes the characters on a car plate based on the given conditions and instructions. The car plate recognition system is installed in many places such as toll gates, parking lots and also entrance of highly secured buildings. Even polices are using this application because they can detect speeding vehicles from distance away. These systems are beneficial because it can automate car park management, improve the security of car park operator and the users as well, eliminate the usage of swipe cards and parking tickets, improve traffic flow during peak hours, detect speeding cars on highways, and detect cars which run over red traffic lights.
Among the car plate recognition systems available worldwide are LPR [1, 4], SIREVIA [2], ALPR [3], VLP [5] and etc. However, these systems are not suitable to be applied in Malaysia due to different styles of car plates endorsed. In Malaysia, the developments of image processing applications for the above said purpose are still inadequate where it is unable to reach the 100% accuracy in recognition eventhough various techniques have been applied. Due to that, it is recommended that research is still conducted for this application because of the importance of car plate recognition.

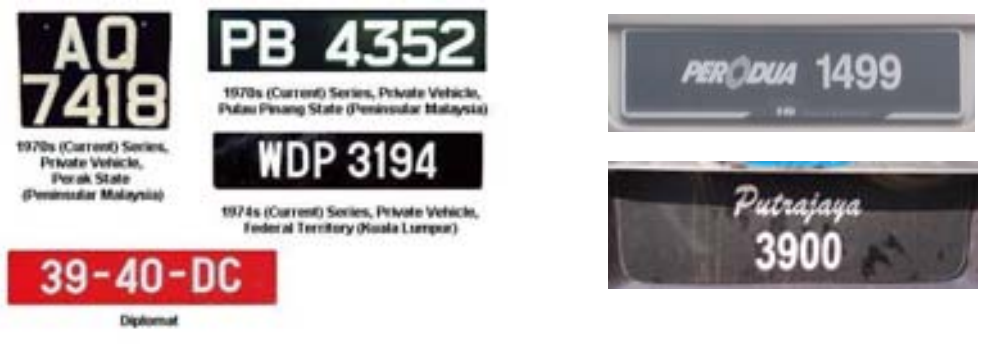

(a) (b)

Fig. 1 (a) Samples of common Malaysian car plates. (b) Samples of special Malaysian car plates with various styles of fonts and sizes

The Road Transport Department of Malaysia has endorsed a specification for car plates that includes the font and size of characters that must be followed by car owners. However, there are cases where this standard is not being followed. Private car owners tend to use various kinds of fonts and sizes for their car plates. Fig. 1 below shows samples of common and special Malaysian car plates. This various fonts and sizes of characters will lead to problems during recognition phase.
One of the factors that contribute to the failure in achieving 100% accuracy in recognition was unable to recognize similar pattern characters such as in the case of recognizing 'B' and '8', 'O' and '0', 'G' and '6' [13], character B or 3 as 8, character 5 as 6, character 6 as G , character A as 4 and character I as 1 [14], character 3 with 8, 4 with A, 8 with B and D with 0 [15] and "8" and "B" or "0" and "D" [16].
Therefore, this paper will focus on conducting an experiment by applying chain codes technique for recognition to be considered as an alternative solution for recognizing characters in Malaysian car plates. Single line Malaysian car plates will be used as testing images.

## 2. Related Works

There are many techniques applied by different researchers in the recognition phase. The techniques are such as multi-methods [6, 8], hybrid methods [7], Hausdorff Distance [9] and minimum distance classifier and neural network method [10]. Due to the different types of characters used in car plates such as the combination of Chinese and Roman characters [6, 7] makes the use of multi-methods or hybrid methods become popular in recognizing car plates. Other reasons why the combination of several techniques becomes a preference are the ability to improve the computation time which is very essential in any real-time car plate recognition system and the ability to shows a high percentage of recognition accuracy.

As proposed by [6], applying multi-methods in performing recognition can increase the correction rate and reduce the computation time and has been approved by a result of 98% accuracy. The multi-methods are the combination of projection sequence feature matching and template matching to recognize Chinese characters and Roman characters in car plate.

Another popular combination of technique is hybrid method which was proposed by [7]. In this research, statistical and structural recognition methods have been applied and the result shows that the method applied is more effective and robust. However, to recognize Chinese character is still a challenge for the researcher.

Performing multi-methods in recognition does not always gives a satisfying result in terms of processing time [8]. The combination of proposed three steps in the recognition approach: the character categorization, topological sorting and self-organizing (SO) recognition only improved the recognition rate but not the time complexity. This was due to the neural-based OCR process running on a sequential computer.

Many researchers applied single technique to perform recognition of characters in car plates. Hausdorff Distance has been chosen as a technique to recognize Thai car plates [9] and proven to achieve high percentage of recognition accuracy. The proposed system was able to perform 92% for valid recognition but needed more research to get high performance in recognizing poor quality images and among groups of similar-pattern characters.

Another research that applied single technique was done by [10]. Malaysian car plates are used as the test subject of car plate recognition. However, two methods of recognition were used to perform the recognition in order to find the best result. The methods are the minimum distance classifier (Method A) and three-layer neural network (Method B). This research however does not shows a very impressive result due to having some kind of 'mustache' portions generated in the thin line formation and less training samples were used.

## 3. Chain Codes

Chain codes are one of the shape representations which are used to represent a boundary by a connected sequence of straight line segments of specified length and direction. This representation is based on 4-connectivity or 8-connectivity of the segments [17]. The direction of each segment is coded by using a numbering scheme as shown in Figure 2 below. Chain codes based from this scheme are known as Freeman chain codes.

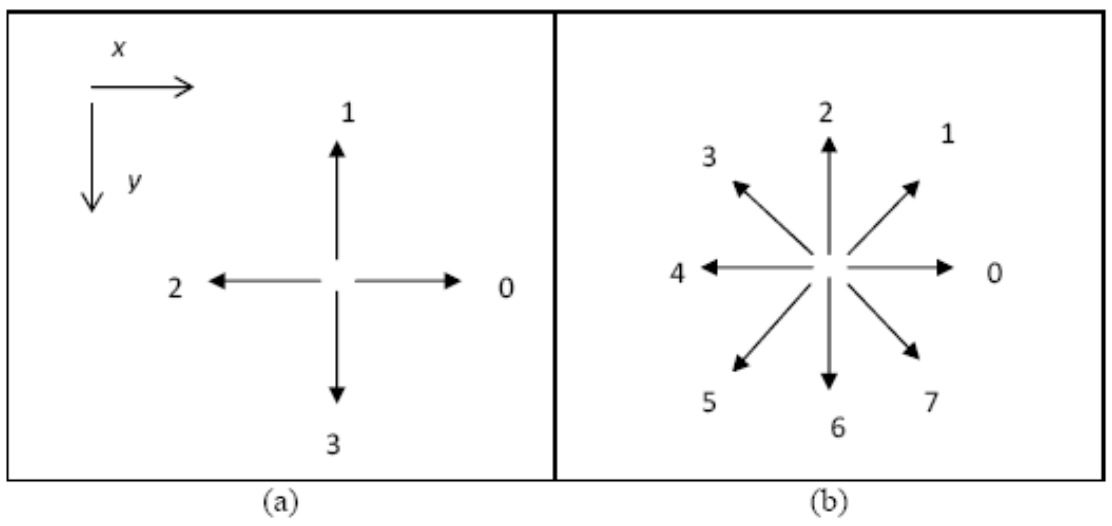

Fig. 2 Direction numbers for (a) 4-directional chain codes, (b) 8-directional chain code

A coding scheme for line structure must satisfy three objectives [18]:

a. It must faithfully preserve the information of interest;
b. It must permit compact storage and convenient for display; and
c. It must facilitate any required processing.

According to [19], chain codes are a linear structure that results from quantization of the trajectory traced by the centers of adjacent boundary elements in an image array. A chain code can be generated by following a boundary of an object in a clockwise direction and assigning a direction to the segments connecting every pair of pixels.

First, we pick a starting pixel location anywhere on the object boundary. Our aim is to find the next pixel in the boundary. There must be an adjoining boundary pixel at one of the eight locations surrounding the current boundary pixel. By looking at each of the eight adjoining pixels, we will find at least one that is also a boundary pixel. Depending on which one it is, we assign a numeric code of between 0 and 7 as already shown in Figure 2. For example, if the pixel found is located at the right of the current location or pixel, a code "0" is assigned. If the pixel found is directly to the upper right, a code "1" is assigned. The process of locating the next boundary pixel and assigning a code is repeated until we came back to our first location or boundary pixel. The result is a list of chain codes showing the direction taken in moving from each boundary pixel to the next. The process of finding the boundary pixel and assigning a code is shown in Figure 3. This method is unacceptable for two main reasons:

a. The resulting chain of codes tends to be quite long
b. Any small disturbances along the boundary due to noise or imperfect segmentation cause changes in the code that may not be related to the shape of the boundary.

Chain codes have been claimed as one of the techniques that are able to recognize characters and digits successfully [20]. This is because of several advantages possessed by this technique as listed by [21]. The first advantage over the representation of a binary object is that the chain codes are a compact representation of a binary object. Second, the chain codes are a translation invariant representation of a binary object. Due to that, it is easier to compare objects using this technique. The third advantage is that the chain code is a complete representation of an object or curve. This means that we can compute any shape feature from the chain codes. According to [22], chain codes provide a lossless compressing and preserving all topological and morphological information which bring out another benefit in terms of speed and effectiveness for the analysis of line patterns.

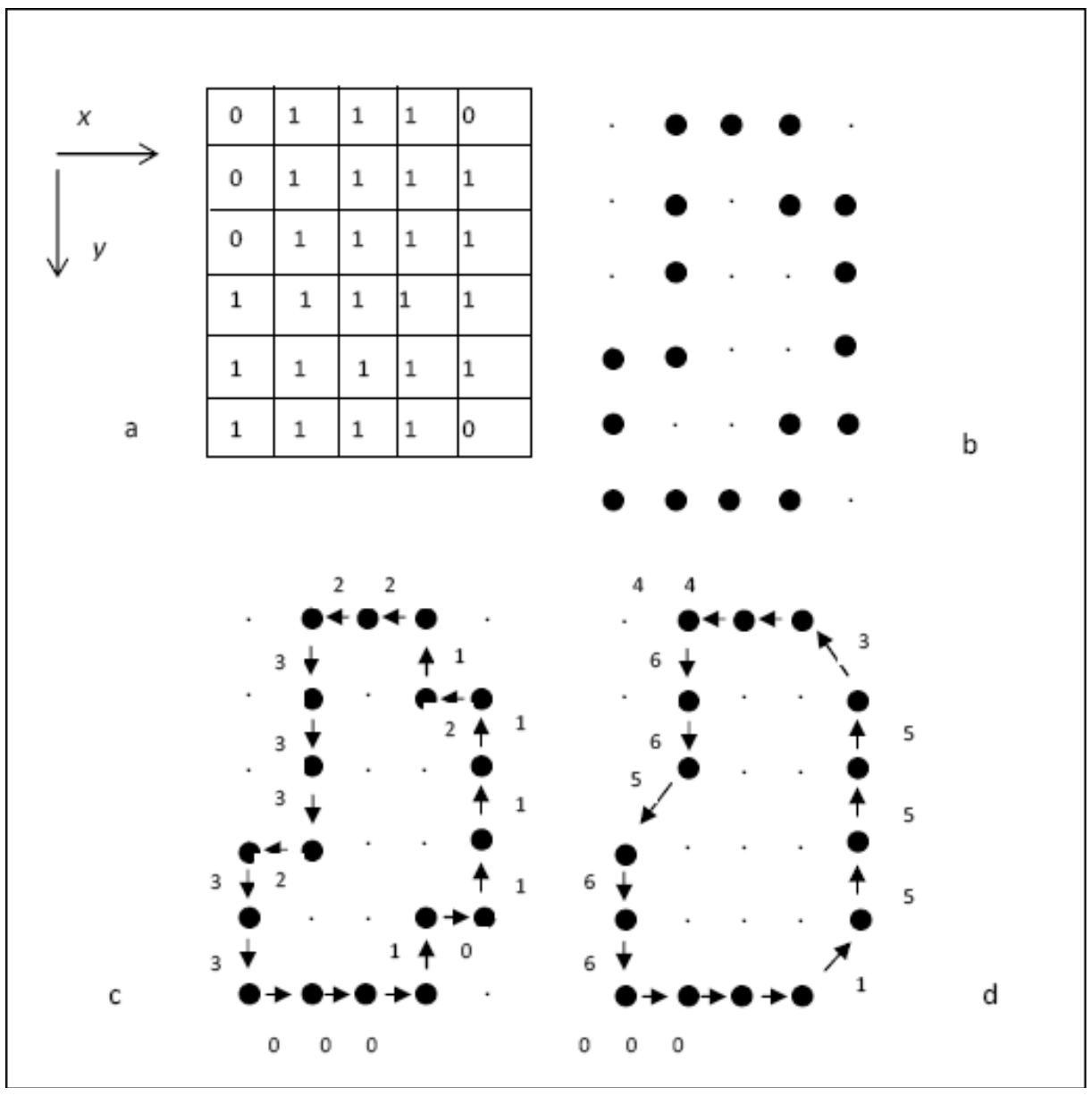

Fig. 3 a & b) A 4-connected object and its boundary; c & d) Obtaining the chain code from the object in (a & b) with (c) for 4-connected and (d) for 8-connected

## 4. Methodology

The Fig. 4 below shows the methodology for this research. There are six phases involved; image acquisition, data definition, image pre-processing, image segmentation, chain code derivation and character recognition. The images of car plates are varied in terms of its sizes and fonts for the testing process (during recognition).

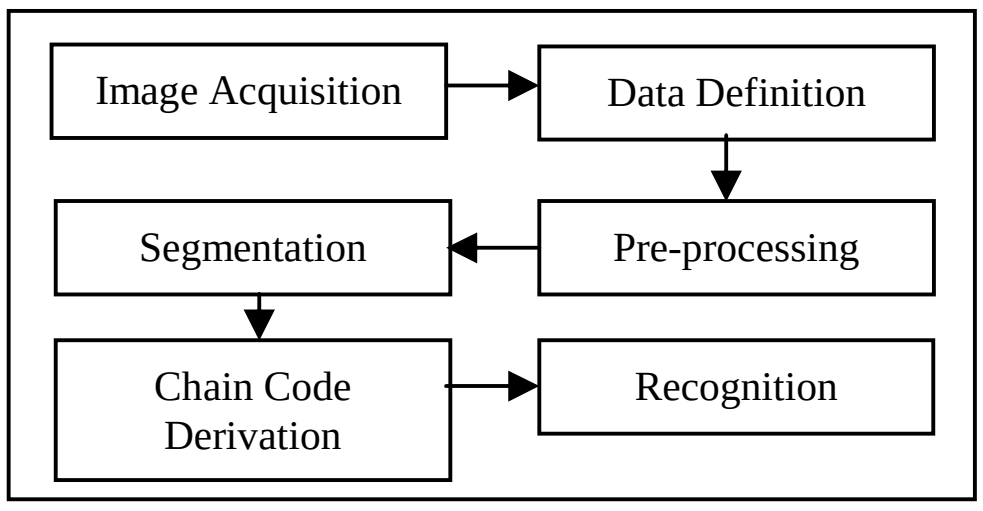

Fig. 4 The proposed methodology

### 4.1 Data Definition

Data that comes from the form of images of car plates need to be identified and analyzed first before tested. The information about types of fonts used need to be gathered since our problem is that the different types of fonts have been used in car plates which lead to problem in recognizing characters by using computer. Therefore, by having this information, all data to be tested will get good results. Based on the observation on collected images of car plates, there are about one to four types of fonts that have been used in car plates (referring to standard car plates). Table 1 below shows types of fonts observed from the collection of images obtained that has been used as raw data.

Table 1 Classification of similar patterns of character '1' taken from four different car plates

| | |
|---|---|
| 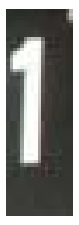 Class 1 | 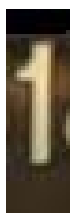 Class 3 |
| 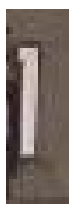 Class 2 | 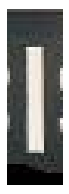 Class 4 |

### 4.2 Preprocessing

Two processes are involved in pre-processing which are threshold and filtering. Images used are gray-scaled images and are converted into binary images which means that every pixel in the image is convert to the binary values (“0” and “1”). Fig. 5 shows image of car plate in the form of gray-scale and after threshold.

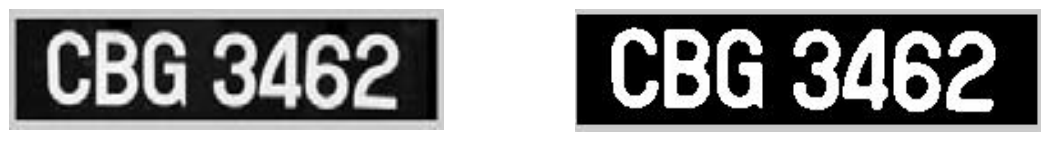

(a) (b)

Fig. 5. (a) Gray-scaled image of car plate. (b) Image after threshold process.

## 4.3 Segmentation

Two processes have been done in this phase; the boundary extraction and segmentation. The boundary image of car plate extraction is done in order to ease the process of deriving the chain codes and illustrated in Fig. 6. The segmentation phase or character isolation takes the region of interest (from the boundary image) and attempts to divide the region into individual characters. To help in detecting the characters, the plate image is divided into seven images where each will contain one isolated character. For the purpose of research, only car plate images which contain 7 characters, with 3 letters at the position of C1, C1 and C3 and 4 numbers at the position N1, N2, N3 and N4 will be used. The segmentation has been done using the pixel count technique first. The connected component labeling technique has been performed for other images which were failed to be segmented using previous technique.

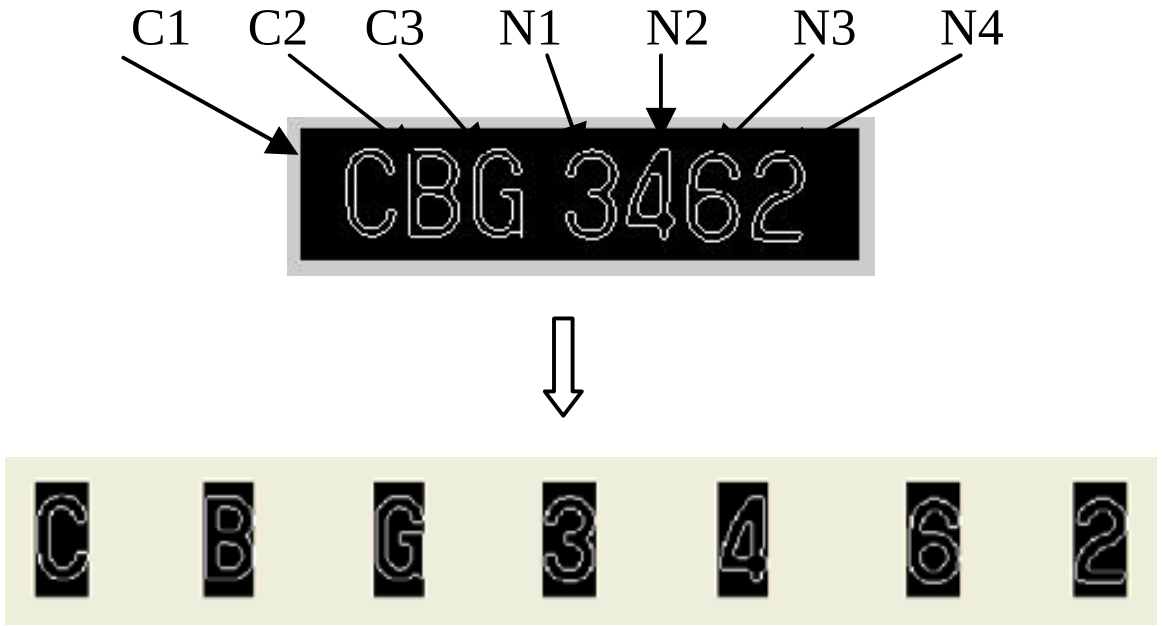

Fig. 6 The boundary image and 7 segmented regions

## 4.4 Chain Code Derivation

This phase is to derive the chain codes for each character or number in the specific region which is the result from image segmentation phase. The algorithm for extracting chain codes for 8-connected boundaries is as follows [23]:

1. Find the pixel in the object that has the leftmost value in the topmost row; call this pixel $P_0$.

2. Define a variable *dir* (for direction), and set it to equal to 7(since $P_0$ is the top-left pixel in the object, the direction to the next pixel must be 7).

3. Traverse the 3x3 neighborhood of the current pixel in a counter-clockwise direction, beginning the search at the pixel in direction dir + 7 (mod 8) *if dir is even* or dir + 6 (mod 8) *if dir is odd.* This will sets the current direction to the first direction counter-clockwise from *dir*:

| *dir* | 0 | 1 | 2 | 3 | 4 | 5 | 6 | 7 |
|---|---|---|---|---|---|---|---|---|
| *dir* + 7 (mod 8) | 7 | 0 | 1 | 2 | 3 | 4 | 5 | 6 |
| *dir* + 6 (mod 8) | 6 | 7 | 0 | 1 | 2 | 3 | 4 | 5 |

4. The first foreground pixel will be the new boundary element. Update *dir.*

5. Stop when the current boundary element $P_n$ is equal to the second element $P_1$ and the previous boundary pixel $P_{n-1}$ is equal to the first boundary element $P_0$.

Fig. 7 below shows the initial location ($P_0$) and the direction derived using the above algorithm while Fig. 8 illustrates the list of chain codes of character ‘C’.

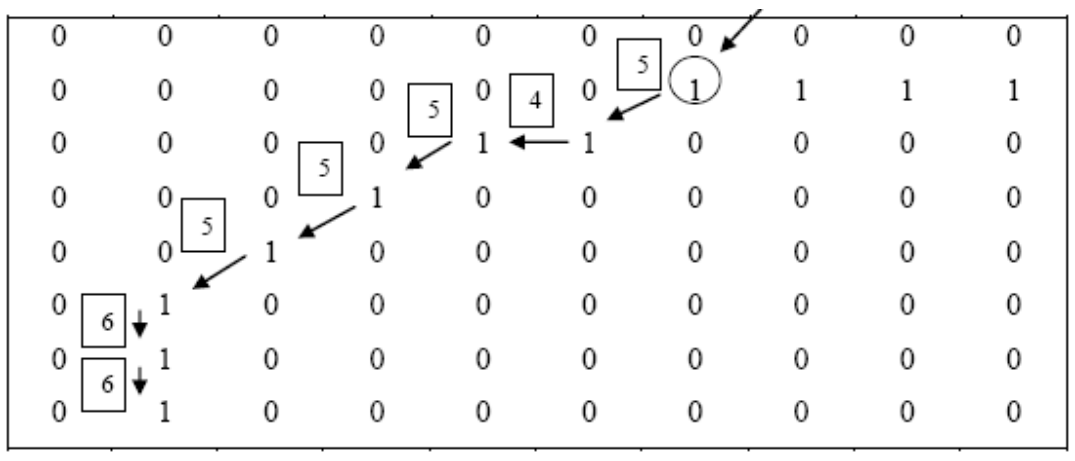

Fig. 7 The initial location, $P_0$ and the direction to derive chain codes.

Columns 1 through 19
5 4 4 5 4 5 4 5 5 6 6 5 6 5 6 6 6 6 6
Columns 20 through 38
6 6 6 6 6 6 6 6 6 6 6 6 6 6 7 6 6 6 6
Columns 39 through 57
6 7 6 7 7 7 7 7 0 0 7 0 0 0 0 0 0 1 0
Columns 58 through 76
1 0 1 1 1 2 1 2 2 1 2 2 2 3 4 4 4 4 5
Columns 77 through 95
6 6 6 5 5 5 5 4 4 4 4 4 3 3 3 2 3 2 2
Columns 96 through 114
2 2 2 2 2 2 2 2 2 2 2 2 2 2 2 2 2 2 2
Columns 115 through 133
2 1 1 1 1 0 0 0 7 0 7 7 7 6 7 6 7 0 0
Columns 134 through 152
0 1 2 2 2 2 3 2 2 3 3 3 3 4 3 4 4 4 3
Columns 153 through 154
4 4

Fig. 8. The chain code extracted from the boundary image of character ‘C’

### 4.5 Recognition

The character recognition has been done by using the list of chain codes derived for each character from the previous phase. It works by calculating the total of each code contained in the list of chain codes. The total of each code is then used as a guide to recognize the characters by matching the chain codes extracted from the previous phase.

## 5. Result and Discussion

An experiment using 110 images of car plates have been done with different types of fonts have been classified into several classes of similar pattern of characters depending on samples of images collected. The results have been observed in two different ways; the recognition accuracy and the computational time. Table 1 shows the result of segmentation and recognition accuracy.

Table 2: Results of segmentation and recognition

| Phase | Technique | Successful | Percentage |
|---|---|---|---|
| Segmentation | Pixel count | 94/110 | 85.45% |
| | Connected component labeling | 105/110 | 95.45% |
| Recognition | FCC | 80/105 | 76.19% |

From the table, we can see that different techniques give different result in terms of the segmentation accuracy. Pixel count technique can only achieved up to 85.45% while connected component labeling is up to 95.45%. Eventhough the rate is quite high, however the best technique should be able to reach almost 100% accuracy. In terms of recognition accuracy rate, using FCC only achieved up to 76.19%. This is considered as low rate since our aim is to achieve almost 100% accuracy.

Recognition time for each character is about 0.01s to 0.07s. and varied with the differentiation in types of fonts but most characters have same recognition time regardless its font types. For example, character ‘1’ has recognition time of only 0.01s for all types of fonts.

From the analysis, the high recognition time for certain font types of characters are due to variation of sizes of fonts used for car plates. Fonts with thin and small size have less recognition time compares to fonts with thick and big size. This factor gives impact for the recognition time because it is related to the number of pixels for all seven characters in a car plate which are used during the processing to calculate the recognition time. As one of the features that are used to recognize objects (for this case, the characters of car plates), the number of pixels do play an important part where logically, an increasing of the number of pixels will increase the recognition time.

As a conclusion, based on the experiment, not all types of fonts used in Malaysian car plates can be recognized by using the Freeman chain codes technique due to its disability to give a high recognition accuracy rate. Eventhough it has an advantage of less recognition time which is very promising however, we do look at the recognition accuracy rate as important as recognition time. To achieve high recognition rate it is suggested that this technique should be applied with other technique which is able to derive the specific or unique features of each character to avoid the error in recognition or to improve the recognition rate.

## References

[1] Xu Jianfeng, Li Shaofa and Chen Zhibin (2003). Color Analysis for Chinese Car Plate Recognition. *Proceedings of the IEEE International Conference on Robotics, Intelligent Systems and Signal Processing*. pp1312 – 1316.

[2] Pedro Barroso, Joaquim Amaral, André Mora, José Manuel Fonseca, Adolfo Steiger-Garção (2004). A Quadtree Based Vehicles Recognition System. *4th WSEAS International Conference on Optics, Photonics, Lasers And Imaging (ICOPLI 2004)*. pp12-16.

[3] Wenjing Jia, Huaifeng Zhang and Xiangjian He, “Region-Based License Plate Detection”, *Journal of network and Computer Applications*, Elsevier, In Press, Corrected Proof, Available online 17 November 2006.

[4] Yo-Ping Huang, Shi-Yong Lai, Wei-Po Chuang, “A Template-Based Model For License Plate Recognition”, Proceedings of the 2004 IEEE International Conference on Networking, Sensing and Control, vol. 2, pp. 737 – 742, 2004.

[5] Tran Duc Duan, Tran Le Hong Du, Tran Vinh Phuoc, Nguyen Viet Hoang, “Building an Automatic Vehicle License-Plate Recognition System”, 3rd International Conference on Computer Science - Research, Innovation & Vision for the Future (RIVF’05), pp. 59-63, 2005.

[6] Qingchuan Tao, Xiaohai He, Daishen Luo and Wei Wu, “A New Car Plate Recognition Method Based On Fuzzy Entropy”, Intelligent Control and Automation, 2004. WCICA 2004. Fifth World Congress on Volume 5, 15-19 June 2004, pp. 4054 - 4056 Vol.5

[7] Xiang Pan, Xiuzi Ye and Sanyuan Zhang, “A Hybrid Method For Robust Car Plate Character Recognition”, Engineering Applications of Artificial Intelligence Volume 18, Issue 8 , December 2005, pp. 963-972

[8] Shyang-Lih Chang, Li-Shien Chen, Yun-Chung Chung, and Sei-Wan Chen, "Automatic License Plate Recognition" Intelligent Transportation Systems, IEEE Transactions on Volumes 5 Issue 1, March 2004 Page(s): 42-53

[9] Juntanasub, R.; Sureerattanan, N., "Car License Plate Recognition through Hausdorff Distance Technique", Tools with Artificial Intelligence, 2005. ICTAI05. 17th IEEE International Conference on Volume, Issue, 14-16 Nov. 2005 Page(s): 5 pp. 647 - -651

[10] M.Fukumi, Y.Takeuchi, H.Fukumoto, Y.Mitsura and M.Khalid, "Neural Network Based Threshold Determination for Malaysia License Plate Character Recognition", Proceedings of 9th International Conference on Mechatronics Technology, vol.1, pp.1-5, 2005
[11] P. Patrick van der Smagt (1990). A Comparative Study Of Neural Network Algorithms Applied To Optical Character Recognition. *Proc. of the 3rd international Conference on Industrial and Engineering Applications of Artificial Intelligence and Expert Systems*, pp1037-1044.
[12] Lucas J. Van and Ben J.H. Verwer (1988). A Contour Processing Method for Fast Binary Neighbourhood Operations. *Pattern Recognition Letters*, 7(1): 27-36.
[13] Cheokman Wu, Lei Chan On, Chan Hon Weng, Tong Sio Kuan and Kengchung Ng (2005). A Macao License Plate Recognition System. *Proceedings of the Fourth International Conference on Machine Learning and Cybernatics*.
[14] Siti Norul Huda Sheikh Abdullah,Marzuki Khalid and Rubiyah Yusof (2006). License Plate Recognition Using Multi-Cluster And Multilayer Neural Networks. *2nd International Conference on Information and Communication Technologies*. pp1818-1823.
[15] Shen-Zheng Wang and Hsi-Jian Lee. (2007). A Cascade Framework For A Real-Time Statistical Plate Recognition System. *IEEE Transactions On Information Forensics And Security*, 2(2).
[16] Qian Gao, Xinnian Wang and Gongfu Xie. License Plate Recognition Based On Prior Knowledge. *Proceedings of the IEEE International Conference on Automation and Logistic.* August 18 – 21. Jinan, China: 2007
[17] Gonzales, R. C and Woods, R. E. (2002). Digital Image Processing.2$^{nd}$ Ed. Upper Saddle River, N. J.: Prentice-Hall, Inc.
[18] Freeman H, Computer Processing of Line-Drawing Images, ACM Computing Surveys, Vol. 6, No.1, 1974, pp57-97
[19] S. Madhvanath, G. Kim and V. Govindaraju, "Chaincode Contour Processing for Handwritten Word Recognition", IEEE Transactions on Pattern Analysis and Machine Intelligence, Vol. 21, No. 9, September 1999
[20] Jahne, B. (2005). *Digital Image Processing*. 6$^{th}$ ed.New York: Springer.
[21] Seul et al (1999). *Practical Algorithms for Image Analysis: Description, Examples and Code*. edisi kebrp. USA: Cambridge University Press.
[22] McAndrew, A. (2004). *Introduction to Digital Image Processing With Matlab*. USA: Thomson Course Technology. :pg353

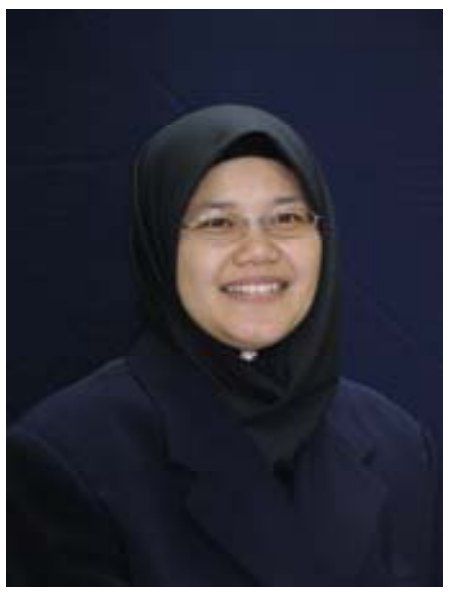

**Nor Amizam Jusoh** received the B.S. degree in Computer Science from University Technology of Malaysia in 2000. Currently, she is pursuing her M.Sc degree at the Faculty of Computer Systems and Software Engineerings, University Malaysia Pahang. She is now a Lecturer at IKIP International College since 2009. Her research interest includes image processing and bioinformatics.

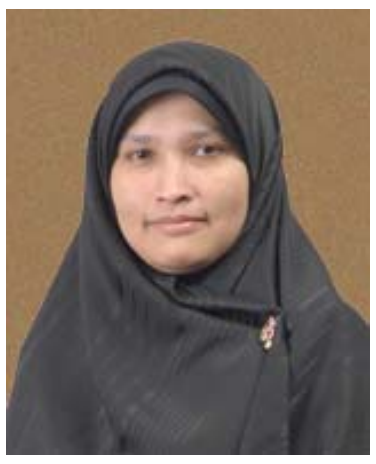

**Jasni Mohamad Zain** received her Bachelor degree in Computer Science from University of Liverpool, England, UK in 1989; PGCE Mathematics from Sheffield Hallam University, England, UK in 1994; M.Ed. degree from Hull University, England, UK in 1998 and PhD from Brunel University, West London, UK in 2005. She currently holds the post as the Dean of Faculty of Computer Systems & Software Engineering, University Malaysia Pahang. She has been actively presenting papers in national and international conferences. Her research interests include Image Processing as well as Data and Network Security.